\documentclass[conference]{IEEEtran}
\usepackage{cite}
\usepackage{amsmath,amssymb,amsfonts}
\usepackage{algorithm}
\usepackage{graphicx}
\usepackage{textcomp}
\usepackage{xcolor}
\usepackage[hyphens]{url}
\usepackage{fancyhdr}
\usepackage{array}
\usepackage{tabularx}
\usepackage{bm}
\usepackage{booktabs}
\usepackage{authblk}
\usepackage[switch]{lineno}
\usepackage{multirow}
\usepackage{adjustbox}
\usepackage{float}
\usepackage{makecell, hhline}

\usepackage{graphicx}
\usepackage{multirow}
\usepackage{adjustbox}
\usepackage{amsmath}
\usepackage{amssymb}
\usepackage{booktabs}

\usepackage{graphicx}
\usepackage{amsmath}
\usepackage{amssymb}
\usepackage{booktabs}
\usepackage[accsupp]{axessibility}

\usepackage{graphicx}
\usepackage{multirow}
\usepackage{adjustbox}

\usepackage{algorithm}
\usepackage{algorithmicx}
\usepackage{algpseudocode}
\usepackage{bm}

\usepackage[pagebackref,breaklinks,colorlinks]{hyperref}

\usepackage[capitalize]{cleveref}
\crefname{section}{Sec.}{Secs.}
\Crefname{section}{Section}{Sections}
\Crefname{table}{Table}{Tables}
\crefname{table}{Tab.}{Tabs.}

\def\BibTeX{{\rm B\kern-.05em{\sc i\kern-.025em b}\kern-.08em
    T\kern-.1667em\lower.7ex\hbox{E}\kern-.125emX}}

\title{PV-VTT: A Privacy-Centric Dataset for Mission-Specific Anomaly Detection and Natural Language Interpretation}
\author[1,*]{Ryozo Masukawa}
\author[1,*]{Sanggeon Yun}
\author[2]{Yoshiki Yamaguchi}
\author[1, $\dag$]{Mohsen Imani}
\affil[1]{University of California, Irvine}
\affil[2]{Shibaura Institute of Technology}
\affil[$*$]{Equal Contribution}
\affil[$\dag$]{Corresponding author, email: m.imani@uci.edu}

\begin{document}
\maketitle
\thispagestyle{plain}
\pagestyle{plain}

\begin{abstract}
    Video crime detection is a significant application of computer vision and artificial intelligence. 
However, existing datasets primarily focus on detecting severe crimes by analyzing entire video clips, often neglecting the precursor activities (i.e., privacy violations) that could potentially prevent these crimes.
To address this limitation, we present \textbf{PV-VTT} (\underline{\textbf{P}}rivacy \underline{\textbf{V}}iolation \underline{\textbf{V}}ideo \underline{\textbf{T}}o \underline{\textbf{T}}ext), a unique multimodal dataset aimed at identifying privacy violations. \textbf{PV-VTT} provides detailed annotations for both video and text in scenarios.
To ensure the privacy of individuals in the videos, we only provide video feature vectors, avoiding the release of any raw video data. This privacy-focused approach allows researchers to use the dataset while protecting participant confidentiality.
Recognizing that privacy violations are often ambiguous and context-dependent, we propose a Graph Neural Network (GNN)-based video description model. Our model generates a GNN-based prompt with image for Large Language Model (LLM), which deliver cost-effective and high-quality video descriptions. By leveraging a single video frame along with relevant text, our method reduces the number of input tokens required, maintaining descriptive quality while optimizing LLM API-usage. Extensive experiments validate the effectiveness and interpretability of our approach in video description tasks and flexibility of our \textbf{PV-VTT} dataset. Dataset available here : \hyperlink{https://ryozomasukawa.github.io/PV-VTT.github.io/}{https://ryozomasukawa.github.io/PV-VTT.github.io/}.
\end{abstract}
\section{Introduction}
\label{sec:intro}
\noindent
Video crime detection is a critical application of computer vision and artificial intelligence. Several datasets have been developed to facilitate this task, with the most widely used being XD-Violence~\cite{Wu2020not}, UCF-Crime~\cite{sultani2018real}, and ShanghaiTech~\cite{liu2018ano_pred}. UCF-Crime and XD-Violence focus on radical and severe crime cases, such as burglary, abuse, and explosions. Conversely, the ShanghaiTech dataset includes a broader range of anomalies, from relatively minor incidents like jumping to serious cases like theft, captured through surveillance cameras. Despite the advancements represented by these datasets, they are typically analyzed at the level of the entire video clip, which means predictions are made only after a crime has already occurred.
This raises a fundamental question: \textit{Can we improve machine learning models to prevent severe crimes before they happen?} To achieve this, there is a need for datasets that can capture preemptive indicators of such crimes.

One significant aspect preceding serious crimes, such as burglary, is \textbf{Privacy Violation}. Potential burglars might engage in preliminary activities like entering private property, taking photographs to identify entry points, or using drones to inspect a building's layout. These preparatory actions involve privacy violations and necessitate automated surveillance to protect potential victims and facilitate swift responses. To date, no datasets have been specifically proposed for the detection of privacy violations. Motivated by this gap, we introduce \textbf{PV-VTT} (\underline{\textbf{P}}rivacy \underline{\textbf{V}}iolation \underline{\textbf{V}}ideo \underline{\textbf{T}}o \underline{\textbf{T}}ext), a new dataset containing video related to various privacy violations frame feature vectors and text descriptions for each video. 

Compared to the radical crime scenarios in existing datasets, privacy violations are often more nuanced and challenging to assess. For instance, 
defining trespassing requires understanding whether the intruder had permission from the property owner, as stipulated by laws such as California State Law \cite{callaw602}. A simplistic model that flags all potential trespassing without contextual understanding risks generating numerous false positives. Thus, models designed to detect privacy violations must be capable of reasoning and providing explanations in natural language to ensure their decisions are both accurate and interpretable.
To date, while video description datasets has been extensively introduced~\cite{7780940, zhou2018towards}, none have concentrated on explaining privacy violations. 
To address this gap, we provide text descriptions for all privacy-related videos in the \textbf{PV-VTT} dataset, which are anonymized using a Large Language Model (LLM)-based semi-automated annotation process, following established methods for video description datasets.
\textbf{PV-VTT} dataset supports benchmarking tasks in both Video Anomaly Recognition (VAR) and video description, representing a fundamental novelty in this field.

Weakly supervised models such as AnomalyCLIP~\cite{zanella2023delving} and MissionGNN~\cite{yun2024missiongnn} have already achieved state-of-the-art results on challenging VAR datasets like XD-Violence. Given that our \textbf{PV-VTT} dataset closely resembles XD-Violence in structure,
we hypothesized that these models, particularly MissionGNN, would achieve high performance in \textbf{PV-VTT} dataset, which is supported by our empirical results. However, existing VAR methods cannot generate descriptions of the detected anomalies. 
To address this, our key contribution is the development of a novel benchmarking method that integrates GNN message passing with LLM prompting for video description.
Unlike earlier methods requiring complex gradient computations in large foundation models~\cite{zhou2022learning}, our approach utilizes message passing capabilities and knowledge graph (KG) integration as a prompt of LLMs. By extracting the top-$k$ keyword edges that most significantly contribute to classification and using them as prompts for LLMs, we significantly reduce the number of tokens needed for API-usage to generate an effective video descriptions using LLMs while maintaining performance comparable to feeding the entire video as a prompt. This reduction in token usage also lowers overall costs, as LLM APIs typically charge based on token usage.

In summary, our primary contributions are:
\begin{itemize}
    \item We introduce \textbf{PV-VTT}, a novel dataset specifically designed for privacy violation detection, enabling research across tasks such as Video Anomaly Recognition (VAR) and video description.
    \item We evaluate several state-of-the-art (SOTA) models on \textbf{PV-VTT}. The experimental results highlight both the challenges and the significance of the dataset.
    \item We propose a cost-effective LLM prompting framework that integrates message passing and knowledge graphs (KGs) into a prompt. This framework demonstrates highly interpretable reasoning capabilities while optimizing LLM API usage.
\end{itemize}

\section{Related Works}
\subsection{VAR and Video Description Datasets}
\subsubsection{Video Anomaly Detection Datasets}
Various video anomaly detection datasets have been introduced in the computer vision research community to date. In this context, we exemplify existing works and elucidate how \textbf{PV-VTT} distinguishes itself from these.

The CUHK Avenue Dataset~\cite{lu2013abnormal}, one of the earliest anomaly detection datasets, comprises 31,000 frames focusing on detecting anomalous behaviors in a subway environment. The ShanghaiTech Campus Dataset~\cite{liu2018ano_pred} captures both normal and anomalous events across 13 different areas on a university campus, with a notable scarcity of normal frames. The UCF-Crime dataset~\cite{sultani2018real} includes 1,900 surveillance videos featuring 13 public safety anomalies, with 800 normal and 810 anomalous videos for training, and 150 normal and 140 anomalous videos for testing. XD-Violence~\cite{Wu2020not} focuses on violence detection, containing 4,754 videos with audio and weak labels, covering six types of violent anomalies. It is split into 3,954 training and 800 testing videos, spanning a total of 217 hours. 

The CHAD dataset~\cite{danesh2023chad} is the most recent contribution to this field. It contains over 1.15 million frames within a single context, accompanied by detailed annotations for detection, tracking, and pose estimation. CHAD provides the largest portion of normal frames among existing datasets. However, its focus is limited to a parking lot scenario, which lacks diversity in background situations.

In contrast to these previous datasets, our \textbf{PV-VTT} dataset is close to XD-violence and UCF-Crime in terms of annotation methods and video diversity. Compared to those 2 datasets, we specialize in cases involving privacy violations and \textbf{PV-VTT} contains a diverse set of labels and includes videos of varying lengths, amounting to approximately 2.3 million frames in total. To the best of our knowledge, \textbf{PV-VTT} is the first to provide both video content and descriptive annotations in this domain.

\subsubsection{Video Description Datasets}
Existing video description datasets can be divided into two primary categories. The first category includes pairwise datasets that pair instructional videos with detailed step-by-step descriptions for each segment~\cite{zhou2018towards, krishna2017dense, huang2020multimodal}. For example, ActivityNet Captions~\cite{krishna2017dense} builds upon the original ActivityNet by providing videos with temporally annotated sentence descriptions. This dataset contains 20,000 videos, each annotated with 100,000 sentences describing various events occurring throughout the video. These annotations vary in duration and may overlap, providing a comprehensive insight into the video content. YouCook2 ~\cite{zhou2018towards} focuses on 2,000 cooking videos featuring 89 recipes sourced from YouTube, while ViTT ~\cite{huang2020multimodal} extends upon YouCook2 with additional instructional videos beyond cooking scenarios.

The second category consists of straightforward video clip-text description pairs ~\cite{chen-dolan-2011-collecting, 7780940}, exemplified by datasets such as MSVD ~\cite{chen-dolan-2011-collecting} and MSR-VTT ~\cite{7780940}. MSVD comprises 1,970 videos featuring single-activity clips, each accompanied by multilingual captions created through crowdsourcing. Conversely, MSR-VTT offers concise descriptions for short video clips spanning 20 categories, totaling 10,000 video clips with 200,000 clip-sentence pairs. \textbf{PV-VTT} adopts a similar clip-text annotation approach but focuses on instances of describing privacy violations.

\subsection{GNN for Capturing Multimodal Information}
Graph Neural Networks (GNNs), such as Graph Convolution Networks (GCNs) \cite{kipf2016semi} and Graph Attention Networks (GATs) \cite{velivckovic2017graph}, are crucial for processing structural data represented by nodes and edges. The message passing network framework \cite{gilmer2017neural} generalizes GNNs, comprising three key functions: message passing, aggregation, and updating. These functions compute messages along edges, aggregate them from neighboring nodes, and update node embeddings based on these aggregations, capturing the graph's underlying structure.
Extending message passing for multi-modal information, recent research developed GNN models for domains like information retrieval \cite{gao2022graph}, Vision Question Answering (VQA), and reasoning \cite{Marino_2021_CVPR, mavromatis2024gnn, Gao_2020_CVPR}, often combining them with knowledge graphs. ConceptNet \cite{speer2017conceptnet}, a broad semantic knowledge graph, is frequently used to build detailed graphs enhancing model reasoning \cite{wang2022vqa, sorokin2018modeling, bosselut2019comet}. We incorporate ConceptNet in our framework for generating automated knowledge graphs as a key structured source. Specifically, VQA-GNN \cite{Wang_2023_ICCV} is a Visual Question Answering model that uses a multimodal GNN to fuse unstructured and structured knowledge, integrating scene and concept graphs via a QA context super node, greatly improving reasoning and surpassing baselines on various VQA benchmarks. MissionGNN~\cite{yun2024missiongnn} is similar to VQA-GNN, but our task, video description, is fundamentally different. Our multi-modal GNN structure uses a goal-oriented Directed Acyclic Graph (DAG). We aim for the GNN-based model to provide detailed explanations for ambiguous privacy violation cases. Methods like GNNExplainer \cite{ying2019gnnexplainer} interpret GNNs by generating masks that highlight crucial subgraphs and node features. This approach uses mean field variational approximation to simplify and clarify visual insights. Unlike GNNExplainer, we employ normalized message weights to generate effective explanations of classification results. Our approach uses a goal-oriented DAG structure, simplifying the explanation process by logically aligning with the classification outcome.
\subsection{Cost-effective LLM usage}
Managing large language models (LLMs) demands significant resources, making their API calls costly, especially for small businesses. For instance, using GPT-3 for customer service costs \$21,000 a month~\cite{gptcost}, which is unaffordable for many. Strategies like FrugalGPT~\cite{chen2023frugalgpt} and Data Shunt~\cite{chen2024data} aim to reduce these costs. FrugalGPT identifies cost-effective LLM combinations, while Data Shunt integrates models to enhance performance and cut costs. However, these methods do not cover video description. GPT-4o~\cite{gpt4o} handles multi-modal data but is expensive for video descriptions due to the vast number of tokens required for each frame. Our novel method proposes a cost-effective video-description approach using single image and graph edges derived from small GNN-model's output.

\begin{figure*}[h!]
    \centering
    \includegraphics[width=0.85\linewidth]{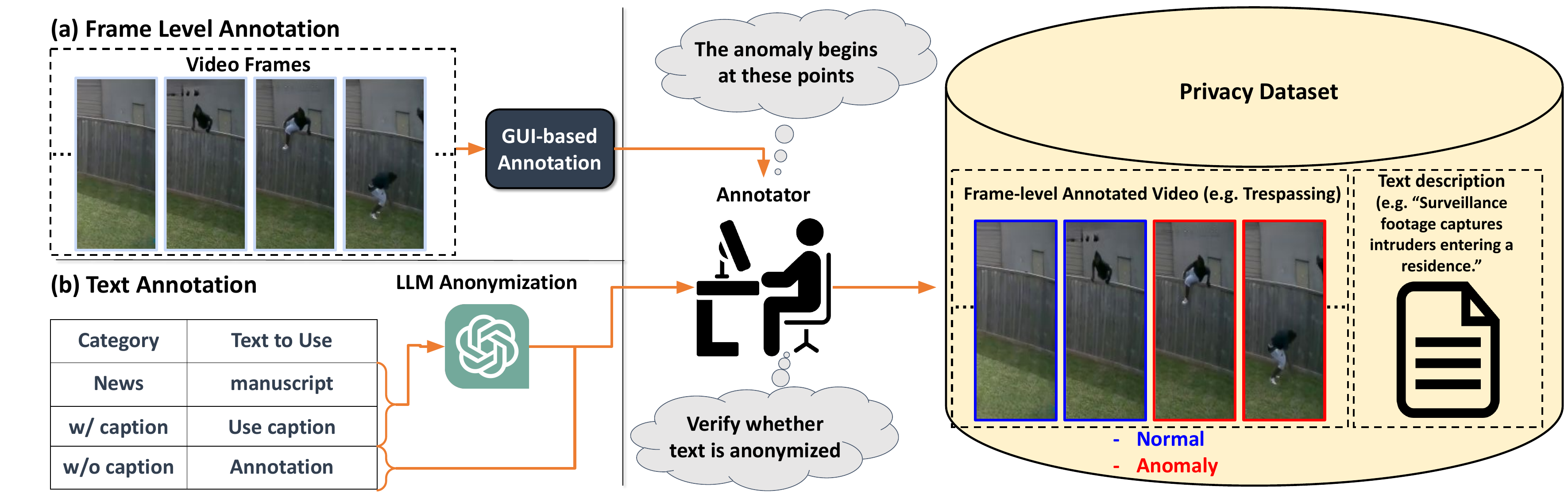}
    \caption{Overview of the privacy video data collection. (a) Frame-level annotation process. (b) Video description generation.}
    \label{fig:data_collection}
\end{figure*}

\begin{figure*}[h!]
    \centering
    \includegraphics[width=\textwidth]{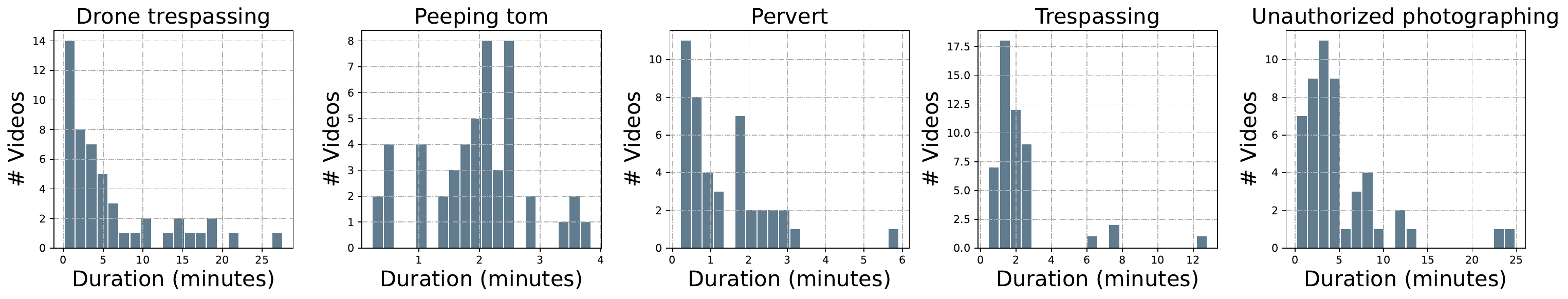}
    \caption{Distribution of video length (minutes)}
    \label{fig:video-duration}
\end{figure*}

\begin{figure}
    \centering
    \includegraphics[width=0.48\textwidth]{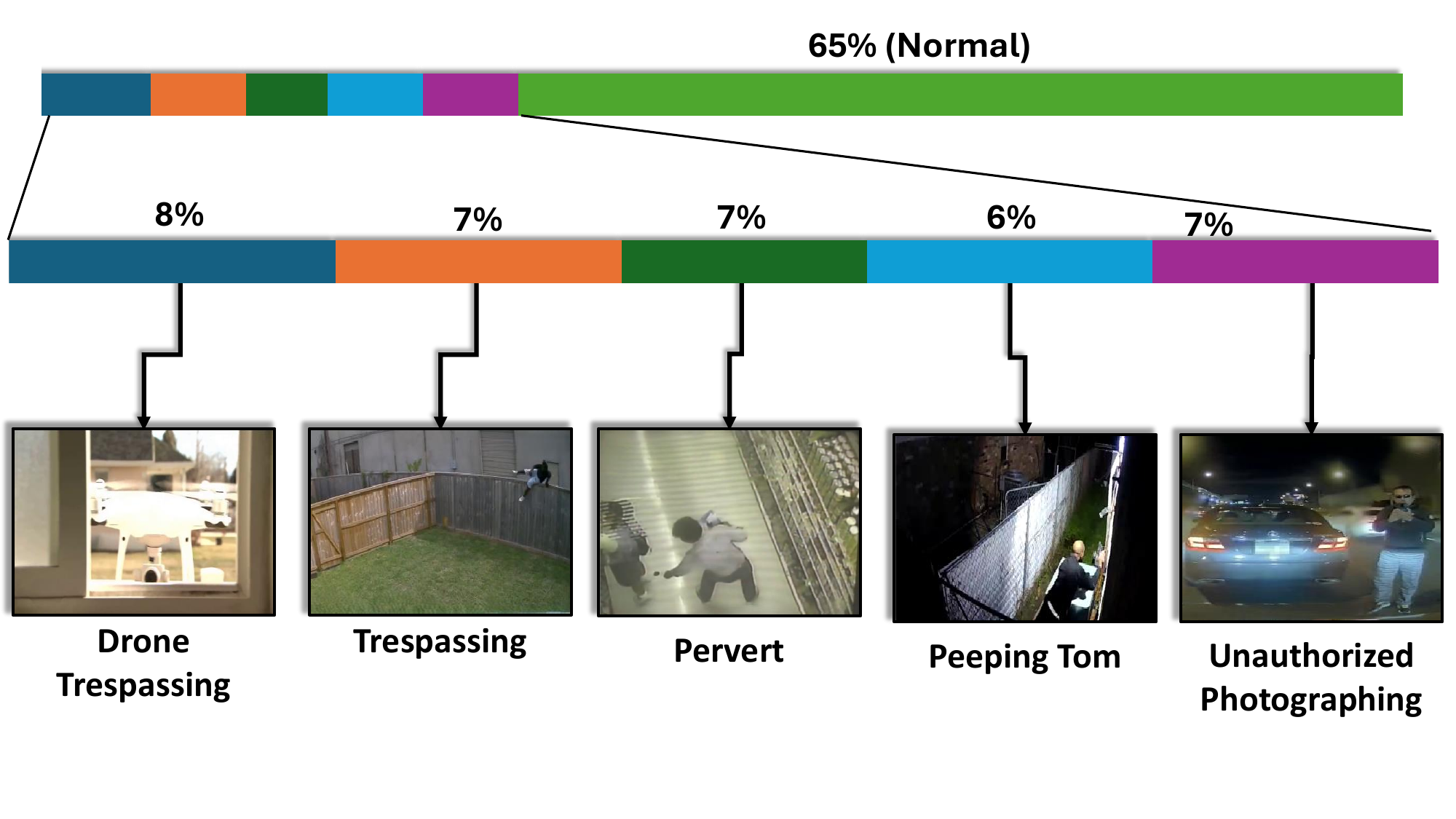}
    \caption{Distribution of Privacy Violations by Case}
    \label{fig:potion}
\end{figure}

\begin{table}[h!]
    \centering
    \caption{Comparison of VAR Video datasets. DivLoc. stands for videos are taken in diverse locations}
    \label{tab:cmp}
    \begin{adjustbox}{width=0.45\textwidth}
    \begin{tabular}{cccccc}
    \toprule
    \textbf{Name} &  \textbf{\#Videos} &  \textbf{Duration} & \textbf{DivLoc.} & \textbf{Crime?} & \textbf{Description?}   \\
    \midrule
    YouCook2 & 2000 & 176h & Yes & No & Yes \\
    UCF-Crime & 1900 & 128h & Yes & Yes & No\\
    XD-violence & 4754 & \textbf{217h} & Yes & Yes & No \\
    CHAD & - & 11h & No & Yes & No \\
    \midrule
    Ours & 702 & 31h &  \textbf{Yes} &  \textbf{Yes} & \textbf{Yes} \\
    \bottomrule
    \end{tabular}%
  \end{adjustbox}%
\end{table}
\section{PV-VTT Dataset}
\subsection{Privacy Violation Categories}
The United States law ~\cite{uslaw} defines privacy invasion  as ``Invasion of privacy involves the infringement upon an individual's protected right to privacy through a variety of intrusive or unwanted actions. Such invasions of privacy can range from physical encroachments onto private property to the wrongful disclosure of confidential information or images." This definition is yet ambiguous and variety of cases can be considered as privacy invasions. Therefore, it is essential to establish a taxonomy for privacy invasions to enable human annotators to classify privacy violations. As a result, we defined five cases, namely \textit{Peeping Tom, Trespassing, Pervert, Drone Trespassing} and \textit{Unauthorized Photographing.} These classes are based on previous political science study on privacy violations~\cite{pyle1982invasion, unhumanrights}. Conversely, \textit{Normal} videos are defined as those that do not contain the activities described previously, and where all individuals depicted have provided consent to be shown.

\subsection{Data Collection and Annotation}
The overall collection method for our proposed dataset is depicted in \autoref{fig:data_collection}. 
We collected a diverse range of privacy invasion videos from video sharing platforms such as YouTube. These videos include footage captured by both fixed surveillance cameras and mobile cameras. Thus, our dataset provide a variety of videos from different backgrounds and camera conditions. Each video was annotated at the frame level by human annotators(\autoref{fig:data_collection} (a)). Specifically, the frame-level annotation indicates that the human annotator watches the video and selects the frames where the behavior that violates privacy both begins and ends. Frames within this annotated range are labeled as positive, while all other frames are labeled as negative.

At the same time, we annotated the text descriptions for videos containing privacy violation incidents using the following method. From the perspective of text description, the videos we gathered can be categorized as follows: 1) \textit{News}: based on news reports; 2) \textit{Video with caption}: which provides a detailed caption about the content of the video on a video-sharing website; and 3) \textit{Video without caption}. For videos in categories 1 and 2, the granularity is so high that the raw data could infringe on the video providers' privacy, as it contains sensitive information such as the names of victims and perpetrators, locations, etc. Therefore, anonymization is essential to make our dataset publicly available. Since manual anonymization by human annotators is too labor-intensive, we employed an LLM-based framework to anonymize the original texts. We used the prompt \textit{``Paraphrase the given text so that the result does not include any information that can identify specific individuals, such as locations or names,"} to obtain anonymized texts. Finally, a human annotator performs a sanity check on each data point to ensure that the LLM has removed all privacy information from the original texts. For videos in category 3, we provided annotations based on handwritten descriptions by human experts. Detailed information is provided in \autoref{fig:data_collection} (b). Note that \textbf{PV-VTT} does not contain description on \textit{Normal} videos because the usage of \textbf{PV-VTT} dataset is solely for enabling video description model to explain privacy violations.
To protect the privacy of individuals and comply with copyright laws, we do not make the actual videos public. Instead, we provide a frame-level embedded vector by ImageBind~\cite{girdhar2023imagebind} without exposing the raw footage. 

\subsection{Dataset Statistics}
Our video dataset includes recordings from both stationary CCTV cameras and mobile cameras operated by witnesses of privacy violation incidents. This combination results in a diverse collection of footage capturing various settings and camera perspectives. The distribution of video lengths and the proportion of each type of privacy violation case are illustrated in \autoref{fig:video-duration} and \autoref{fig:potion}, respectively. To highlight the distinctive features and uniqueness of our dataset, we provide a comparative analysis in \autoref{tab:cmp}, where we evaluate it against existing datasets from both the VAR and video description perspectives. Notably, our dataset is novel in its combination of crime captioning with video data.

In summary, \textbf{PV-VTT} offers several strengths. First, it features meticulous frame-level annotation, enabling researchers to train models in both fully supervised and weakly supervised methods. Second, it includes a wide variety of scenarios, providing a rich diversity in the dataset. Third, it contains detailed textual explanations that allow future models trained on our dataset to understand and reason why privacy violations occur.

\section{Methodology}

\begin{figure*}[h!]
    \centering
    \includegraphics[width=1.0\textwidth]{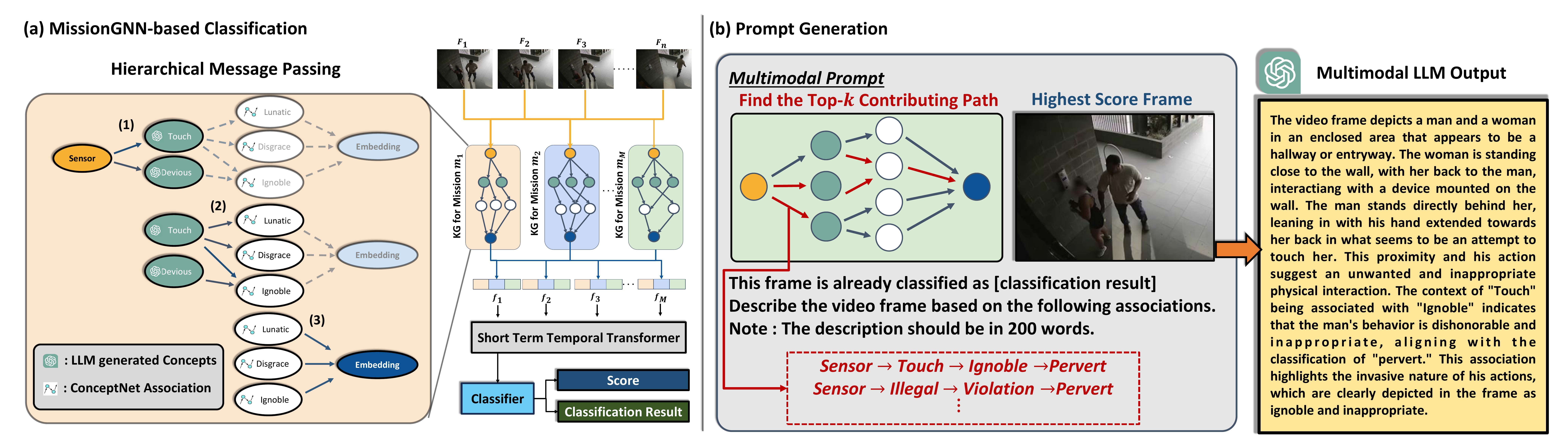}
    \caption{(a) An overview of Mission-specific Knowledge Graph Generation and Video Classification: Messages are always passed from hierarchcally (1) sensor data nodes to LLM-generated key concept nodes, (2) key concept nodes to ConceptNet association nodes, and (3) association nodes to the final embedding node. (b) Framework to generate LLM prompt from the pretrained MissionGNN model.}
    \label{fig:kg-gen}
\end{figure*}

To detect and explain privacy violations in \textbf{PV-VTT}, we propose a novel framework that integrates GNN with the visual prompting capabilities of LLMs, leveraging MissionGNN~\cite{yun2024missiongnn}. Our framework utilizes only a single video frame, equivalent to an image, as its prompt, creating a cost-effective approach that enables the LLM to generate comprehensive video descriptions.

\subsection{Mission-specific Knowledge Graph Genetation}
First, the mission-specific KG generation framework is used for generating KG that can extract useful information from a given frame image (\autoref{fig:kg-gen} (a)). In VAR, each mission-specific KG indicates a KG that contains structured knowledge regarding the corresponding privacy violation case. The first step is to obtain related vocabularies which we call \textbf{Key Concepts} for each privacy violation cases using GPT-4o~\cite{gpt4o} using a prompt \textit{``List up} $V(\in \mathbb{N})$  \textit{typical vocabularies to represent [violation name] case? Note: Everything should be in a single word."} After obtaining key concepts, for each concept, we gain a related word using ConceptNet~\cite{speer2017conceptnet} ``relatedTo" relationship and construct edges from key concepts to related vocabularies (e.g. ``ogle" relatedTo ``amorous" in \textit{Pervert} class). On top of the mission-specific knowledge graph, we place a sensor node whose node embedding vector contains sensory information, such as images encoded by joint-embedding models like ImageBind~\cite{girdhar2023imagebind}, and projects directed edges for each key concept node. Finally, all related concept nodes project directed edges to an embedding node that aggregates all messages passed from sensor nodes to key concept and related concept nodes.

The motivation of our knowledge graph design is to allow GNN to pass interpretable messages from multi-modal information. By utilizing multimodal embedding models such as ImageBind, information from all nodes can be embedded into the same vector space.

\subsection{Hierarchical GNN}
After generating $M$ $(\in \mathbb{N})$ privacy violation KGs ($G_{m_{i}}$ $(1 \leq i \leq M$) for each case, we train a hierarchical GNN video anomaly recognition model that classifies which privacy violation occurs in the given video (\autoref{fig:kg-gen} (a)).

GNNs capture relational information using feature vectors for each node, and now we have a KG that connects both embedded sensor node feature \( x^{(0)}_{s, m_i} \) by the multimodal model's image encoder \( \mathcal{E}_I \) and texts from each embedded associated concept nodes \( x^{(0)}_{c, m_i} \) (\( c \in \{ \text{Generated vocabulary set for mission } m_i \} \)) by the text encoder \( \mathcal{E}_T \) in the first layer of GNN projected in the same vector space as follows:

\begin{equation}
    \bm{x}^{(0)}_{s, m_i} = \mathcal{E}_I(F_{t}),\hspace{3mm}\bm{x}^{(0)}_{c, m_i} = \mathcal{E}_T(c)
\end{equation}

where $F_t$ is a video frame at timestamp $t$.

By utilizing simple multi-layer perceptron (MLP) layer, the node feature is embedded to smaller dimensions to effectively represent relationships between nodes of $G_{m_{i}}$:

\begin{equation}
    \label{eq:mlp}
    \bm{x}^{(l)}_{m_{i}} = \phi^{(l)}_{m_{i}}(\bm{x}^{(l-1)}_{m_{i}}) = W^{(l)}_{m_{i}} \bm{x}^{(l-1)_{m_{i}}} + \bm{b}^{(l)}_{m_{i}}
\end{equation}
where $W^{(l)}_{m_{i}}$ denotes trainable weight parameter matrix and $\bm{b}^{(l)}_{m_{i}}$ indicates bias at layer $l$ ($1\leq l \leq L$, $L \in \mathbb{N})$.

The key idea in our framework is a \textbf{Hierarchical Message Passing} layer that selectively passes Graph Neural Network (GNN) messages through distinct levels of the knowledge graph (KG) hierarchy. Specifically, we structure our KG into three hierarchical levels: (1) sensor data nodes to LLM-generated key concept nodes, (2) key concept nodes to ConceptNet association nodes, and (3) association nodes to the final embedding node. 

In traditional GNNs, such as the GCN, message passing involves aggregating information from all neighboring nodes indiscriminately. This approach effectively allows each node to represent the overall features of the graph's geometrical structure. However, in this GNN framework, message passing is constrained to follow the hierarchical levels in our KG, which is a DAG with a clear objective of reaching the embedding node. 

This hierarchical message passing mechanism is crucial for two reasons: First, it respects the directed nature of our KG, ensuring that information flows in a manner consistent with the graph's intended structure. Second, it enables more targeted and efficient aggregation of information from various modalities by specifying the exact types of messages to be received and processed at each level. This design leads to more interpretable and goal-oriented embeddings that are particularly useful for complex, multi-modal knowledge representation tasks.

Hierarchical Message Passing can be described as:

\begin{equation}
    \label{eq:msg}
    \bm{x}^{(l)}_{v} =
    \frac{1}{|\mathcal{N}^{(h-1)}(v)|} \sum_{u\in \mathcal{N}^{(h-1)}(v)} \phi^{(l)}(\bm{x}^{(l-1)}_{v}\cdot \bm{x}^{(l-1)}_{u})  
\end{equation}

where \( \bm{x}^{(l)}_{v} \) represents the feature vector of node \( v \) in the current hierarchy \( h \), \( \mathcal{N}^{(h-1)}(v) \) denotes a set of neighbors of node \( v \) in the preceding hierarchy \( h-1 \). Same MLP function \( \phi^{(l)} \) in \autoref{eq:mlp} is applied to the element-wise product of the feature vectors \( \bm{x}^{(l-1)}_{v} \) and \( \bm{x}^{(l-1)}_{u} \).
For simplicity, mission-specific notations are omitted in \autoref{eq:msg}.

This approach allows each node \( v \) to aggregate information from its neighbors \( u \) in the preceding hierarchy, capturing the hierarchical structure of our knowledge graph, where each layer corresponds to a different level of abstraction or type of information. As a result, final embedding node feature $\bm{x}^{(L)}_{\text{emb}, m_i}$ for each mission $m_i$ can contain all the track of association from sensor to key and associated concept nodes. All mission KG embbedding are combined into a single vector as follows:

\begin{equation}
    \bm{f}^{(t)} = [\bm{x}^{(L)}_{emb, m_{1}}, \bm{x}^{(L)}_{emb, m_{1}},...,\bm{x}^{(L)}_{emb, m_{M}}],
\end{equation}
where ``\textit{emb}" indicates final embedding node. 
By extracting the feature vector for each frame \( F_t \), we define \( X_t \), a sequence of tokens at timestamp $t$, as 

\begin{equation*}
    X_t = \{\bm{f}^{(t-A+1)}, \bm{f}^{(t-A+2)}, \ldots, \bm{f}^{(t)}\},
\end{equation*}

where \( \bm{f}^{(t)} \) represents the feature vector of the frame at timestamp \( t \). The ability to process sequences of varying lengths is essential due to the heterogeneous nature of video frame counts in video detection models. To address this variability, we define the sequence length \( A \), which specifies the number of frames preceding the current frame at time \( t \) that the model will process. This approach enables the model to consistently analyze and make decisions based on a standardized subset of frames, thus ensuring uniformity in the decision-making process across videos with differing frame counts.
By feeding this sequence to Transformer encoder layer $\mathcal{T}$ and following MLP layer, we gain the output classification result $\hat{\bm{y}}$ as follows:

\begin{equation}
    \hat{\bm{y}} = Softmax(MLP(\mathcal{T}(X_{t}))).
\end{equation}

By training through cross-entropy loss, smoothing loss, and decaying threshold-based anomaly localization in MissionGNN, we gain an optimized GNN video anonmaly detector that specifies on VAR task on \textbf{PV-VTT} dataset.
\subsection{Prompt Generation}
\begin{algorithm}
\caption{Impactive Reasoning Path Finding Algorithm}\label{algorithm:1}
\hspace*{\algorithmicindent} \textbf{Input:}\\
\hspace*{\algorithmicindent}\hspace*{\algorithmicindent}$v$: Currently searching node\\
\hspace*{\algorithmicindent}\hspace*{\algorithmicindent}$G(V, E)$: Undirected acyclic graph consisted of vertex set $V$ and edge set $E$\\
\hspace*{\algorithmicindent}\hspace*{\algorithmicindent}$\{w_{uv}\}_{u,v\in V\times V}$: Hierarchical importance of the edges\\
\hspace*{\algorithmicindent}\hspace*{\algorithmicindent}$P$: Total impact of the previous path\\
\hspace*{\algorithmicindent}\textbf{Output:}\\
\hspace*{\algorithmicindent}\hspace*{\algorithmicindent}$\mathcal{P}$: Set of top $k$ path
\begin{algorithmic}[1]
\State $\mathcal{P} \leftarrow \text{Empty Max Heap}$
\State $\mathcal{N} \leftarrow \{\}$
\Procedure{ReasoningPathFinding}{$v$, $G$, $\{w_{uv}\}$, $P$}
    \State $\mathcal{N} \leftarrow \mathcal{N}\cup\{v\}$
    \State $C \leftarrow 0$
    \For{$u\in \{t | (v, t) \in V\}$} \Comment{Can use adjacency list}
        \State $ReasoningPathFinding(u, G, \{w_{uv}\}, P + w_{vu})$
        \State $C \leftarrow C + 1$
    \EndFor
    \If{$C = 0$} \Comment{Reached the embedding node}
        \State $\mathcal{P}.push(\{P, \mathcal{N}\})$
        \If{$\mathcal{P}.length > k$}
            \State $\mathcal{P}.pop()$
        \EndIf
    \EndIf
    \State $\mathcal{N} \leftarrow \mathcal{N}-\{v\}$
\EndProcedure
\end{algorithmic}
\end{algorithm}

After training, we freeze our hierarchical GNN model to examine the contributions of individual edges in the network. This analysis allows us to understand how each edge influences the model's decision-making process. The visual description and detailed algorithm for prompt generation is shown in \autoref{fig:kg-gen} (b) and Algorithm \autoref{algorithm:1}, respectively.

To quantify the impact of each edge, we utilize the norm of the each message passing through that edge. For an edge connecting node \( u \) to node \( v \), let \( \bm{m}_{uv} \) represent the message vector. The impact of this edge is given by the norm of \( \bm{m}_{ij} \), typically calculated using the Euclidean norm:
\[
\|\bm{m}_{uv}\| = \left( \sum_{k} |m_{uv,k}|^2 \right)^{\frac{1}{2}}
\]

By calculating the norm for each edge and averaging it across different hierarchical levels, we construct a weighted Directed Acyclic Graph (DAG). This DAG captures the hierarchical importance of the edges:
\[
w_{uv} = \frac{ \|\bm{m}_{uv}\|}{\sum_{u' \in V^{(h-1)}, v' \in V^{(h)}} \|\bm{m}_{u'v'}\|}
\]
where \( w_{uv} \) denotes the weight of the edge from the node $u(\in V^{(h-1)})$ to the node $v(\in V^{(h)})$ and $V^{(h)}$ indicates a set of nodes in hierarchy $h$.

To identify the most significant paths in contributing to the final embeddings, we compute the total impact for each path in the DAG. For a given path \( P \), the cumulative impact is the sum of the weights of its constituent edges:
\[
\text{Impact}(P) = \sum_{(i,j) \in P} w_{ij}
\]
We then select the top-\( k \) paths with the highest cumulative impacts, as these paths are considered the most influential in the decision-making process (\textbf{Reasoning Path}s).

As described in the red rectangular section of \autoref{fig:kg-gen} (b), our proposed prompt to the LLM includes not only a single video frame, an image chosen by the highest classification score and the template-based prompt, but also the reasonning paths, which contain the natural language reasoning steps derived from our knowledge graph. Consequently, this approach enhances the prompt's contextual relevance.
In traditional methods~\cite{alayrac2022flamingo}, video frames are passed to vision-language models by splitting them into patches, where each patch is tokenized (i.e., converted into a format the model can process, often resulting in a large number of tokens). This process can be inefficient, as it requires processing an enormous number of tokens to handle consecutive video frames. In contrast, our method leverages the structural insights provided by the GNN to produce meaningful video descriptions based on learned representations, reducing the need to tokenize multiple video frames, which significantly decreases the overall token count.

\section{Experiments}
\begin{table*}[h!]
    \centering
    \caption{VAR performance on \textbf{PV-VTT} Dataset}
    \label{tab:cmp_var}
    \begin{adjustbox}{width=\textwidth}
    \begin{tabular}{cccccccc}
    \toprule
    \multirow{2}{*}{\textbf{Method}} & \multicolumn{6}{c}{\textbf{AUC by Class}} & \multirow{2}{*}{\textbf{mAUC}}\\ 
    \cmidrule(lr){2-7} 
     & Normal & Peeping Tom & Pervert & Trespassing & Drone Trespassing & Unauthorized Photographing & \\
    \midrule

    ImageBind + PCA + RF & \textbf{94.06} & 87.25 & 80.42 & 86.09 & \textbf{94.51} & \textbf{99.59} & 89.57 \\ 
    ImageBind + MLP & 74.23 & 97.02 & 94.40 & 92.94 & 91.40 & 90.42 & 93.24 \\ 

    \midrule
    \textbf{MissionGNN}~\cite{yun2024missiongnn} & 74.95 & \textbf{97.43} & \textbf{97.38} & \textbf{93.27} & 93.29  & 91.69  & \textbf{94.61} \\ 
    \bottomrule
    \end{tabular}%
  \end{adjustbox}%
\end{table*}

\subsection{Implementation Details}
For our multimodal embedding approach, we deployed the pre-trained ImageBind huge model. Additionally, for generating mission-specific knowledge graphs (KG), we utilized GPT-4o alongside ConceptNet 5. The optimizer we employ is AdamW~\cite{loshchilov2017decoupled}, configured with a learning rate of $10^{-5}$, a weight decay of $1.0$, $\beta_1 = 0.9$, $\beta_2 = 0.999$, and $\epsilon = 10^{-8}$. We set the decay threshold $\alpha_d$ to $0.9999$. For the hierarchical GNN model, we maintain a uniform dimensionality $D_{m_i, l} = 8$. For the short-term temporal model, we use an inner dimensionality of 128, complemented by 8 attention heads. The model undergoes training for 3,000 steps, with each mini-batch consisting of 128 samples. For all video and text generation tasks, we used GPT-4o.
\subsection{VAR Quality}

In this section, we conduct a rigorous evaluation of the performance of MissionGNN model using the \textbf{PV-VTT} dataset. This evaluation is critical because the effectiveness of our proposed LLM prompt-generation process is inherently dependent on the classification accuracy of MissionGNN. Specifically, if MissionGNN demonstrates inaccuracies in the VAR task when applied to the \textbf{PV-VTT} dataset, the explanations generated as a result of misclassification would lack validity and coherence.

\noindent \textbf{Baselines:} As \textbf{PV-VTT} only make the ImageBind frame-level feature vectors publicly available, future researchers will be expected to train models using these feature values for evaluating the quality of VAR models. To ensure a fair comparison, we evaluate two baseline approaches: a Random Forest (RF) classifier that utilizes ImageBind feature vectors processed with Principal Component Analysis (PCA), and a simple Multi-Layer Perceptron (MLP) classifier. These baseline approaches are compared with our MissionGNN-based method.

\noindent \textbf{Evaluation Metrics:} 
To evaluate Video Anomaly Recognition (VAR), we employ methods consistent with previous research~\cite{Wu2020not, sultani2018real, zanella2023delving}. Specifically, we assess the area under the receiver operating characteristic curve (AUC-ROC) at the frame level within a multi-class classification context. This metric is chosen because it is independent of thresholds, making it particularly suitable for detection tasks; a higher AUC signifies better differentiation between normal and abnormal events. For each class, we calculate the AUC by designating the frames of that class as positives and frames of other classes as negatives. The mean AUC (mAUC) is then computed across all anomalous classes. 

As illustrated in \autoref{tab:cmp_var}, MissionGNN consistently outperforms or matches the performance of existing baselines across all anomalous classes. The superior mAUC achieved by MissionGNN underscores its enhanced ability to classify anomalous cases with greater stability and robustness compared to other methods. These findings validate MissionGNN as the most effective approach fora VAR task using \textbf{PV-VTT} dataset.

\subsection{Evaluation on Video Captioning Quality}

To evaluate our MissionGNN-based video description, we used text descriptions from the \textbf{PV-VTT} dataset as our ground truth. For a fair comparison, we utilized the pretrained MissionGNN VAR method to provide all classification results on the test set of \textbf{PV-VTT}. These classification results were included in the prompts for baseline LLM video captioning, replicating real-world scenarios where our proposed MissionGNN-based video captioning method generates and utilizes prompts.

\noindent \textbf{Baselines:} For comparison of the video description quality, we used GPT4o without any labels but input video file and a simple prompt \textit{"Describe the video"} which we call \textit{GPT-4o (w/o label)}. In addition, we give the template-based prompt that fills the classification result of MissionGNN and video\textit{GPT4o(w/ label)}, and finally ours \textit{GPT4o (+ MissionGNN prompting)}.

\noindent \textbf{Evaluation metrics:} Following widely-used previous video captioning research~\cite{Ko_2023_CVPR, zhou2018towards, 7780940}, to evaluate the video captioning of our proposed MissionGNN prompt generation, we use the same metrics such as BLEU~\cite{papineni2002bleu}, ROUGE-L~\cite{lin2004rouge}, METEOR~\cite{banerjee2005meteor}, and  
BertScore~\cite{zhang2019bertscore}.
We also conducted a human evaluation on a subset of \textbf{PV-VTT}, asking a third-party research group different from us to assess whether the generated descriptions from our proposed method have similar quality with those from the baseline \textit{GPT-4o (w/ label)}, which processes all video frames with high accuracy but at a higher cost due to token usage, and our proposed method, which uses MissionGNN with a single video frame, offering a more cost-effective solution. 

Results in \autoref{tab:cmp_desc} indicate that our approach produces video descriptions of similar quality to the baseline, with the majority of evaluators agreeing with the quality.

\begin{table}[h!]
    \centering
    \caption{Video Caption Generation Quality by Methods}
    \label{tab:cmp_desc}
    \begin{adjustbox}{width=0.48\textwidth}
    \begin{tabular}{cccccccc}
    \toprule
    \multirow{3}{*}{\textbf{Method}} &  \multirow{3}{*}{\textbf{BLEU-4}} &  \multirow{3}{*}{\textbf{ROUGE-L}} & \multirow{3}{*}{\textbf{METEOR}} &  \multirow{3}{*}{\textbf{BERTScore}} & \multirow{3}{*}{\textbf{\#Tokens}} & \textbf{HumanEval }  \\ 
    &&&&&& \textbf{(Percentage of approval} \\
    &&&&&& \textbf{from human evaluators}) \\
    \midrule
    GPT-4o (w/o label) & 24.88  & 11.22 & 17.06 & 84.41 & 9785 & - \\ 
    GPT-4o (w/ label) & 23.00  &  9.82 & 17.40 & 84.57 & 9400 & -\\ 
    \midrule
    Ours & 23.42 & 10.27 & 16.25 & 83.68 & \textbf{1123} & \textbf{80\%} \\
    \bottomrule
    \end{tabular}%
  \end{adjustbox}%
\end{table}

\subsection{Evaluation on cost-effectiveness for API usage}
\begin{figure}[h!]
    \centering
    \includegraphics[width=0.5\textwidth]{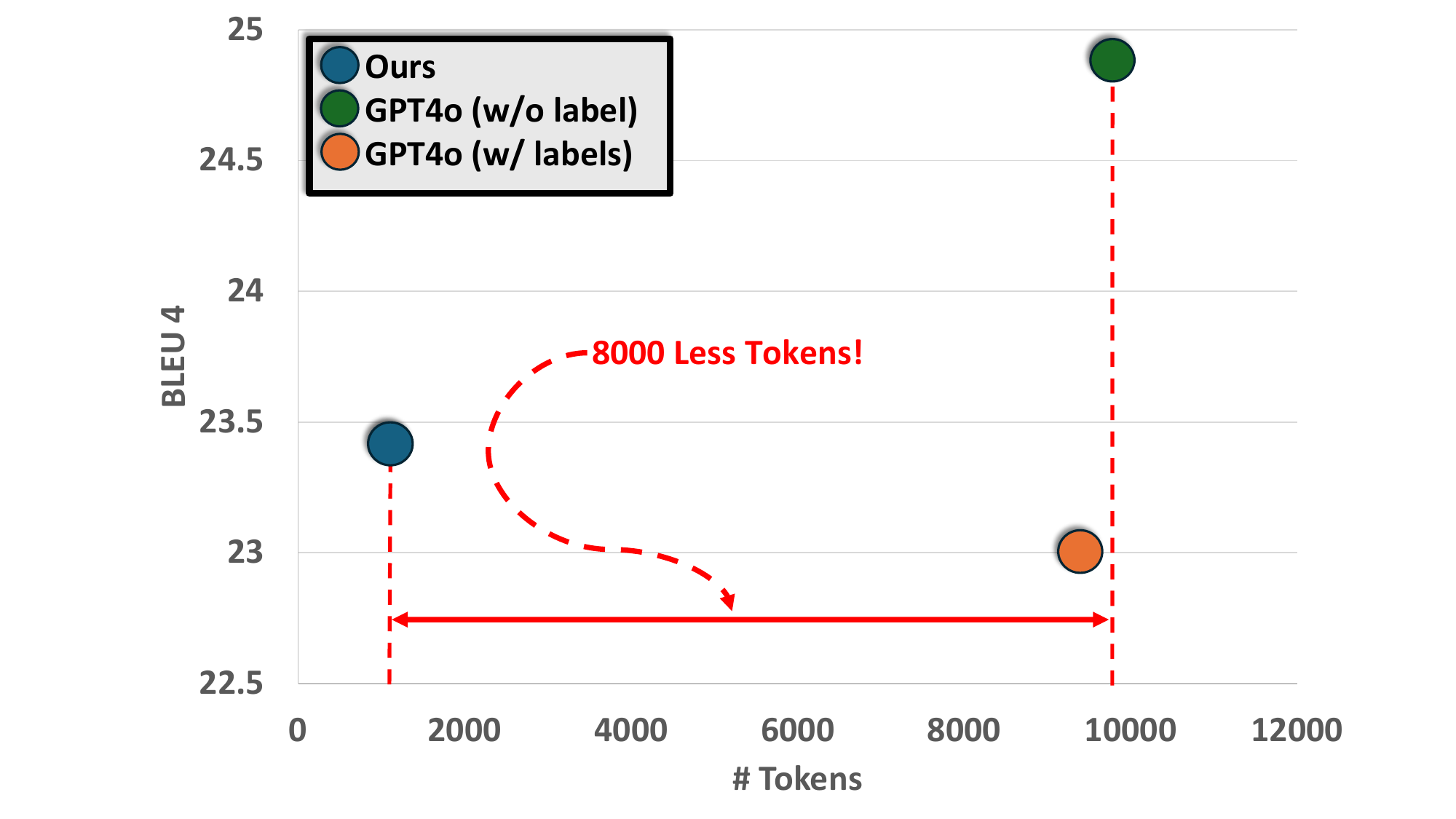}
    \caption{Relationship between Quality and API-usage cost}
    \label{fig:trade-off}
\end{figure}

As the majority of LLM-providing services, such as GPT-4, estimate their charges based on API usage, which correlates to the number of tokens in the API call, we calculated the average number of tokens needed to generate video captions. We then compared how our proposed method cost-effectively reduces API usage.
As shown in \autoref{fig:trade-off} and \autoref{tab:cmp_desc}, our proposed method significantly reduces the number of tokens required for video description, averaging 8,000 fewer tokens than traditional video-based approaches, while achieving superior performance. This efficiency results in almost equivalent quality descriptions with fewer resources. In contrast, using API-based LLMs like GPT-4o to process all video frames without optimization leads to inaccurate results and prohibitively high costs. Applying such naive implementations on a large scale, such as across numerous security cameras, could financially cripple a security company due to the steep API expenses. Our approach mitigates these issues by conserving computational resources and maintaining high output quality, making it highly suitable for extensive deployments. A crucial component of our method is effective knowledge graph (KG) generation, which only requires the usage of LLM API only the initial step. Despite the initial investment for using LLM API to generate mission-specific KG, its benefits are clear: it provides a robust framework for long-term operational sustainability and financial stability. Therefore, investing in thorough KG-generation processes is essential for ensuring the efficiency and viability of large-scale implementations.

\section{Conclusions}
In this paper, we introduced \textbf{PV-VTT}, a novel multimodal privacy dataset that integrates frame-level annotated videos with corresponding text descriptions. To utilize \textbf{PV-VTT} effectively, we proposed a hierarchical GNN-based video captioning technique. Our experiments confirmed the effectiveness of these methods, showing potential for improved multimodal understanding and privacy-sensitive applications. However, some text descriptions in \textbf{PV-VTT} were derived from news manuscripts, containing context not always inferable from the videos, raising concerns about their suitability as ground truth. Future research should consider more precise data curation approaches where detailed, contextually relevant annotations align closely with the video content. Our goal is to advance machine learning for the early detection of nuanced potential criminal activities, contributing to crime prevention. We believe \textbf{PV-VTT} and our methods represent a significant step toward this goal and hope they inspire further research to develop robust systems enhancing public safety and security.

\section*{Acknowledgements}
This work was supported in part by the DARPA Young Faculty Award, the National Science Foundation (NSF) under Grants \#2127780, \#2319198, \#2321840, \#2312517, and \#2235472, the Semiconductor Research Corporation (SRC), the Office of Naval Research through the Young Investigator Program Award, and Grants \#N00014-21-1-2225 and \#N00014-22-1-2067, and Army Research Office Grant \#W911NF2410360. Additionally, support was provided by the Air Force Office of Scientific Research under Award \#FA9550-22-1-0253, along with generous gifts from Xilinx and Cisco.

I would like to thank Wenjun Huang and other members of Bio-Inspired Architecture and Systems Laboratory at the University of California, Irvine for their invaluable assistance with the rebuttal process and for their support in conducting the human evaluation, which greatly contributed to the clarity and rigor of this work.

\bibliographystyle{IEEEtranS}
\bibliography{egbib}

\begin{thebibliography}{10}\itemsep=-1pt

\bibitem{alayrac2022flamingo}
Jean-Baptiste Alayrac, Jeff Donahue, Pauline Luc, Antoine Miech, Iain Barr, Yana Hasson, Karel Lenc, Arthur Mensch, Katherine Millican, Malcolm Reynolds, et~al.
\newblock Flamingo: a visual language model for few-shot learning.
\newblock {\em Advances in neural information processing systems}, 35:23716--23736, 2022.

\bibitem{banerjee2005meteor}
Satanjeev Banerjee and Alon Lavie.
\newblock Meteor: An automatic metric for mt evaluation with improved correlation with human judgments.
\newblock In {\em Proceedings of the acl workshop on intrinsic and extrinsic evaluation measures for machine translation and/or summarization}, pages 65--72, 2005.

\bibitem{bosselut2019comet}
Antoine Bosselut, Hannah Rashkin, Maarten Sap, Chaitanya Malaviya, Asli Celikyilmaz, and Yejin Choi.
\newblock Comet: Commonsense transformers for automatic knowledge graph construction.
\newblock {\em arXiv preprint arXiv:1906.05317}, 2019.

\bibitem{chen-dolan-2011-collecting}
David Chen and William Dolan.
\newblock Collecting highly parallel data for paraphrase evaluation.
\newblock In Dekang Lin, Yuji Matsumoto, and Rada Mihalcea, editors, {\em Proceedings of the 49th Annual Meeting of the Association for Computational Linguistics: Human Language Technologies}, pages 190--200, Portland, Oregon, USA, June 2011. Association for Computational Linguistics.

\bibitem{chen2024data}
Dong Chen, Yueting Zhuang, Shuo Zhang, Jinfeng Liu, Su Dong, and Siliang Tang.
\newblock Data shunt: Collaboration of small and large models for lower costs and better performance.
\newblock In {\em Proceedings of the AAAI Conference on Artificial Intelligence}, volume~38, pages 11249--11257, 2024.

\bibitem{chen2023frugalgpt}
Lingjiao Chen, Matei Zaharia, and James Zou.
\newblock Frugalgpt: How to use large language models while reducing cost and improving performance.
\newblock {\em arXiv preprint arXiv:2305.05176}, 2023.

\bibitem{danesh2023chad}
Armin Danesh~Pazho, Ghazal Alinezhad~Noghre, Babak Rahimi~Ardabili, Christopher Neff, and Hamed Tabkhi.
\newblock Chad: Charlotte anomaly dataset.
\newblock In {\em Scandinavian Conference on Image Analysis}, pages 50--66. Springer, 2023.

\bibitem{gao2022graph}
Chen Gao, Xiang Wang, Xiangnan He, and Yong Li.
\newblock Graph neural networks for recommender system.
\newblock In {\em Proceedings of the Fifteenth ACM International Conference on Web Search and Data Mining}, pages 1623--1625, 2022.

\bibitem{Gao_2020_CVPR}
Difei Gao, Ke Li, Ruiping Wang, Shiguang Shan, and Xilin Chen.
\newblock Multi-modal graph neural network for joint reasoning on vision and scene text.
\newblock In {\em Proceedings of the IEEE/CVF Conference on Computer Vision and Pattern Recognition (CVPR)}, June 2020.

\bibitem{gilmer2017neural}
Justin Gilmer, Samuel~S Schoenholz, Patrick~F Riley, Oriol Vinyals, and George~E Dahl.
\newblock Neural message passing for quantum chemistry.
\newblock In {\em International conference on machine learning}, pages 1263--1272. PMLR, 2017.

\bibitem{girdhar2023imagebind}
Rohit Girdhar, Alaaeldin El-Nouby, Zhuang Liu, Mannat Singh, Kalyan~Vasudev Alwala, Armand Joulin, and Ishan Misra.
\newblock Imagebind: One embedding space to bind them all.
\newblock In {\em Proceedings of the IEEE/CVF Conference on Computer Vision and Pattern Recognition}, pages 15180--15190, 2023.

\bibitem{huang2020multimodal}
Gabriel Huang, Bo Pang, Zhenhai Zhu, Clara Rivera, and Radu Soricut.
\newblock Multimodal pretraining for dense video captioning.
\newblock {\em arXiv preprint arXiv:2011.11760}, 2020.

\bibitem{unhumanrights}
The United Nations General~Assembly in~Paris~on 10~December~1948.
\newblock Universal declaration of human rights, 1948.
\newblock Accessed on June 10th, 2024.

\bibitem{callaw602}
California~Legislative Information.
\newblock California state law section 602.8, 2023.
\newblock Accessed on June 8th, 2024.

\bibitem{kipf2016semi}
Thomas~N Kipf and Max Welling.
\newblock Semi-supervised classification with graph convolutional networks.
\newblock {\em arXiv preprint arXiv:1609.02907}, 2016.

\bibitem{Ko_2023_CVPR}
Dohwan Ko, Joonmyung Choi, Hyeong~Kyu Choi, Kyoung-Woon On, Byungseok Roh, and Hyunwoo~J. Kim.
\newblock Meltr: Meta loss transformer for learning to fine-tune video foundation models.
\newblock In {\em Proceedings of the IEEE/CVF Conference on Computer Vision and Pattern Recognition (CVPR)}, pages 20105--20115, June 2023.

\bibitem{krishna2017dense}
Ranjay Krishna, Kenji Hata, Frederic Ren, Li Fei-Fei, and Juan Carlos~Niebles.
\newblock Dense-captioning events in videos.
\newblock In {\em Proceedings of the IEEE international conference on computer vision}, pages 706--715, 2017.

\bibitem{lin2004rouge}
Chin-Yew Lin.
\newblock Rouge: A package for automatic evaluation of summaries.
\newblock In {\em Text summarization branches out}, pages 74--81, 2004.

\bibitem{liu2018ano_pred}
W. Liu, D.~Lian W.~Luo, and S. Gao.
\newblock Future frame prediction for anomaly detection -- a new baseline.
\newblock In {\em 2018 IEEE Conference on Computer Vision and Pattern Recognition (CVPR)}, 2018.

\bibitem{loshchilov2017decoupled}
Ilya Loshchilov and Frank Hutter.
\newblock Decoupled weight decay regularization.
\newblock {\em arXiv preprint arXiv:1711.05101}, 2017.

\bibitem{lu2013abnormal}
Cewu Lu, Jianping Shi, and Jiaya Jia.
\newblock Abnormal event detection at 150 fps in matlab.
\newblock In {\em Proceedings of the IEEE international conference on computer vision}, pages 2720--2727, 2013.

\bibitem{Marino_2021_CVPR}
Kenneth Marino, Xinlei Chen, Devi Parikh, Abhinav Gupta, and Marcus Rohrbach.
\newblock Krisp: Integrating implicit and symbolic knowledge for open-domain knowledge-based vqa.
\newblock In {\em Proceedings of the IEEE/CVF Conference on Computer Vision and Pattern Recognition (CVPR)}, pages 14111--14121, June 2021.

\bibitem{mavromatis2024gnn}
Costas Mavromatis and George Karypis.
\newblock Gnn-rag: Graph neural retrieval for large language model reasoning.
\newblock {\em arXiv preprint arXiv:2405.20139}, 2024.

\bibitem{gpt4o}
OpenAI.
\newblock Hello gpt-4o, 2024.
\newblock Accessed on June 8th, 2024.

\bibitem{papineni2002bleu}
Kishore Papineni, Salim Roukos, Todd Ward, and Wei-Jing Zhu.
\newblock Bleu: a method for automatic evaluation of machine translation.
\newblock In {\em Proceedings of the 40th annual meeting of the Association for Computational Linguistics}, pages 311--318, 2002.

\bibitem{pyle1982invasion}
Christopher~H Pyle.
\newblock The invasion of privacy.
\newblock {\em Proceedings of the Academy of Political Science}, 34(4):131--142, 1982.

\bibitem{gptcost}
Slowik and Kaiser.
\newblock How much does it cost to use gpt models? gpt-3 pricing explained, 2023.
\newblock Accessed on June 10th, 2024.

\bibitem{sorokin2018modeling}
Daniil Sorokin and Iryna Gurevych.
\newblock Modeling semantics with gated graph neural networks for knowledge base question answering.
\newblock {\em arXiv preprint arXiv:1808.04126}, 2018.

\bibitem{speer2017conceptnet}
Robyn Speer, Joshua Chin, and Catherine Havasi.
\newblock Conceptnet 5.5: An open multilingual graph of general knowledge.
\newblock In {\em Proceedings of the AAAI conference on artificial intelligence}, volume~31, 2017.

\bibitem{sultani2018real}
Waqas Sultani, Chen Chen, and Mubarak Shah.
\newblock Real-world anomaly detection in surveillance videos.
\newblock In {\em Proceedings of the IEEE conference on computer vision and pattern recognition}, pages 6479--6488, 2018.

\bibitem{velivckovic2017graph}
Petar Veli{\v{c}}kovi{\'c}, Guillem Cucurull, Arantxa Casanova, Adriana Romero, Pietro Lio, and Yoshua Bengio.
\newblock Graph attention networks.
\newblock {\em arXiv preprint arXiv:1710.10903}, 2017.

\bibitem{wang2022vqa}
Yanan Wang, Michihiro Yasunaga, Hongyu Ren, Shinya Wada, and Jure Leskovec.
\newblock Vqa-gnn: Reasoning with multimodal semantic graph for visual question answering.
\newblock {\em arXiv preprint arXiv:2205.11501}, 2022.

\bibitem{Wang_2023_ICCV}
Yanan Wang, Michihiro Yasunaga, Hongyu Ren, Shinya Wada, and Jure Leskovec.
\newblock Vqa-gnn: Reasoning with multimodal knowledge via graph neural networks for visual question answering.
\newblock In {\em Proceedings of the IEEE/CVF International Conference on Computer Vision (ICCV)}, pages 21582--21592, October 2023.

\bibitem{uslaw}
Cornel University Law~School Wex Definitions~Team.
\newblock Invasion of privacy, 2023.
\newblock Accessed on June 10th, 2024.

\bibitem{Wu2020not}
Peng Wu, jing Liu, Yujia Shi, Yujia Sun, Fangtao Shao, Zhaoyang Wu, and Zhiwei Yang.
\newblock Not only look, but also listen: Learning multimodal violence detection under weak supervision.
\newblock In {\em European Conference on Computer Vision (ECCV)}, 2020.

\bibitem{7780940}
Jun Xu, Tao Mei, Ting Yao, and Yong Rui.
\newblock Msr-vtt: A large video description dataset for bridging video and language.
\newblock In {\em 2016 IEEE Conference on Computer Vision and Pattern Recognition (CVPR)}, pages 5288--5296, 2016.

\bibitem{ying2019gnnexplainer}
Zhitao Ying, Dylan Bourgeois, Jiaxuan You, Marinka Zitnik, and Jure Leskovec.
\newblock Gnnexplainer: Generating explanations for graph neural networks.
\newblock {\em Advances in neural information processing systems}, 32, 2019.

\bibitem{yun2024missiongnn}
Sanggeon Yun, Ryozo Masukawa, Minhyoung Na, and Mohsen Imani.
\newblock Missiongnn: Hierarchical multimodal gnn-based weakly supervised video anomaly recognition with mission-specific knowledge graph generation.
\newblock {\em arXiv preprint arXiv:2406.18815}, 2024.

\bibitem{zanella2023delving}
Luca Zanella, Benedetta Liberatori, Willi Menapace, Fabio Poiesi, Yiming Wang, and Elisa Ricci.
\newblock Delving into clip latent space for video anomaly recognition.
\newblock {\em arXiv preprint arXiv:2310.02835}, 2023.

\bibitem{zhang2019bertscore}
Tianyi Zhang, Varsha Kishore, Felix Wu, Kilian~Q Weinberger, and Yoav Artzi.
\newblock Bertscore: Evaluating text generation with bert.
\newblock {\em arXiv preprint arXiv:1904.09675}, 2019.

\bibitem{zhou2022learning}
Kaiyang Zhou, Jingkang Yang, Chen~Change Loy, and Ziwei Liu.
\newblock Learning to prompt for vision-language models.
\newblock {\em International Journal of Computer Vision}, 130(9):2337--2348, 2022.

\bibitem{zhou2018towards}
Luowei Zhou, Chenliang Xu, and Jason Corso.
\newblock Towards automatic learning of procedures from web instructional videos.
\newblock In {\em Proceedings of the AAAI Conference on Artificial Intelligence}, volume~32, 2018.

\end{thebibliography}

\end{document}